\documentclass[letterpaper]{article} 
\usepackage{aaai23}  
\usepackage{times}  
\usepackage{helvet}  
\usepackage{courier}  
\usepackage[hyphens]{url}  
\usepackage{graphicx} 
\usepackage{natbib}  
\usepackage{caption} 
\usepackage{algorithmic}
\usepackage{algorithm}
\usepackage{soul}
\usepackage[hidelinks]{hyperref}
\usepackage{url} 
\usepackage{bm} 
\usepackage{multirow}
\usepackage{array}
\usepackage{amsfonts}
\usepackage{amsmath}
\usepackage{mathtools}
\usepackage{tabularx}
\usepackage{booktabs}
\usepackage{subfigure}

\usepackage{newfloat}
\usepackage{listings}
\DeclareCaptionStyle{ruled}{labelfont=normalfont,labelsep=colon,strut=off} 
\floatstyle{ruled}
\newfloat{listing}{tb}{lst}{}
\floatname{listing}{Listing}
\title{Scaling Up Dynamic Graph Representation Learning \\via Spiking Neural Networks}

\author {
    Jintang Li\textsuperscript{\rm 1},
    Zhouxin Yu\textsuperscript{\rm 1},
    Zulun Zhu\textsuperscript{\rm 2},
    Liang Chen\textsuperscript{\rm 1}\footnote{Corresponding author.},\\
    Qi Yu\textsuperscript{\rm 2},
    Zibin Zheng\textsuperscript{\rm 1},
    Sheng Tian\textsuperscript{\rm 3},
    Ruofan Wu\textsuperscript{\rm 3},
    Changhua Meng\textsuperscript{\rm 3}
}

\affiliations {
    \textsuperscript{\rm 1}Sun Yat-sen University\\
    \textsuperscript{\rm 2}Rochester Institute of Technology\\
    \textsuperscript{\rm 3}Ant Group\\
    \{lijt55, yuzhx6\}@mail2.sysu.edu, \{chenliang6, zhzibin\}@mail.sysu.edu,\\ zz2023@g.rit.edu, qi.yu@rit.edu, \{tiansheng.ts,ruofan.wrf,changhua.mch\}@antgroup.com
}

\usepackage{bibentry}

\begin{document}

\maketitle

\begin{abstract}

    Recent years have seen a surge in research on dynamic graph representation learning, which aims to model temporal graphs that are dynamic and evolving constantly over time. However, current work typically models graph dynamics with recurrent neural networks (RNNs), making them suffer seriously from computation and memory overheads on large temporal graphs. So far, scalability of dynamic graph representation learning on large temporal graphs remains one of the major challenges. In this paper, we present a scalable framework, namely SpikeNet, to efficiently capture the temporal and structural patterns of temporal graphs. We explore a new direction in that we can capture the evolving dynamics of temporal graphs with spiking neural networks (SNNs) instead of RNNs. As a low-power alternative to RNNs, SNNs explicitly model graph dynamics as spike trains of neuron populations and enable spike-based propagation in an efficient way. Experiments on three large real-world temporal graph datasets demonstrate that SpikeNet outperforms strong baselines on the temporal node classification task with lower computational costs. Particularly, SpikeNet generalizes to a large temporal graph (2.7M nodes and 13.9M edges) with significantly fewer parameters and computation overheads. Our code is publicly available at \url{https://github.com/EdisonLeeeee/SpikeNet}.
\end{abstract}

\section{Introduction}
\begin{figure}[t]
    \centering
    \includegraphics[width=\columnwidth]{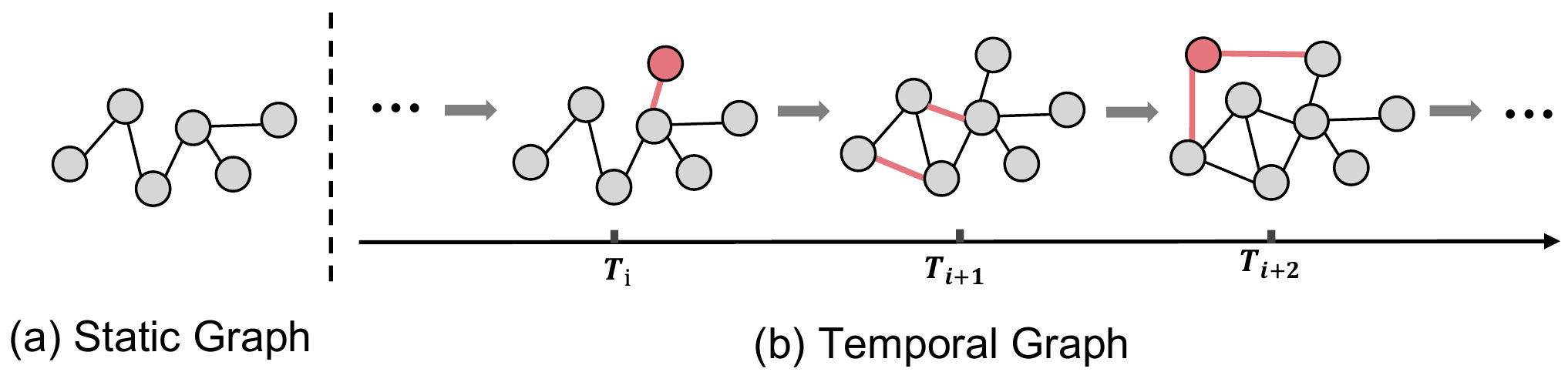}
    \caption{An illustrative example of {(a) static graph} and {(b) temporal graph}. A temporal graph is typically represented as a series of graph snapshots, with network structure constantly evolving over time.}
    \label{fig:comparison}
\end{figure}

A graph is comprised of nodes and edges connected together to model structures and relationships of objects in various scientific and commercial fields~\cite{ogb,chen2020phishing}. It is highly expressive and capable of representing complex structures that are difficult to model, with prominent examples including social networks~\cite{liu2020modelling}, molecules graphs~\cite{zhou2020graph}, and transaction networks~\cite{chen2020phishing}.  In practice, graphs are often dynamic, meaning that nodes, edges, and attributes may evolve constantly over time. Such graphs are typically referred to as \textit{temporal graphs}, in contrast to \textit{static graphs} where nodes and edges remain fixed over time~\cite{dynamic_graph_survey}. Figure~\ref{fig:comparison} is an illustration of static and temporal graphs. The temporal graph is usually represented as a sequence of graph snapshots, while the static graph can be seen as a single (static) observation of the temporal graph.

Over the past few years, graph neural networks (GNNs) have emerged as highly successful tools for learning graph-structured data. The advances in GNNs have led to new state-of-the-art results in numerous graph-based learning tasks~\cite{HamiltonYL17,chen2021understanding}. However, GNNs have been primarily developed for dealing with static graphs while ignoring that the graph itself is dynamically evolving over time. So far, several efforts have been made to generalize current GNNs to temporal graphs by additionally considering the time dimension~\cite{dynamic_graph_survey}. The most established solution for modeling temporal graphs is by extending sequence models to graph data. In this vein, recurrent neural networks (RNNs)~\cite{cho2014learning} are prominent methods that provide a natural choice to capture the temporal information for these evolving graphs~\cite{STAR,EvolveGCN,GAEN,JODIE}. In addition, self-attention~\cite{tgat}, random walk~\cite{DBLP:journals/corr/abs-1901-01346,CAW,DBLP:journals/corr/abs-2108-08754}, and temporal point process~\cite{MMDNE,HTNE,DynamicTriad} are some other approaches to learn structural and temporal patterns simultaneously.

While current approaches have achieved promising results, there are some fundamental challenges when switching from static to temporal graphs. (i) \textit{Scalability}: current research is mostly dominated by RNN-based methods, which require a large number of memory cells to store temporal contextual information~\cite{SakSB14} and large amounts of training data to outperform even static methods. As a result, they suffer from high overheads and scale poorly as the number of time steps increases, especially for large graphs. (ii) \textit{Inductive capability}:
an inductive approach can generate predictions quickly for unseen nodes or links in an evolving graph (e.g., new uses and social behaviors in a social network), which is essential for temporal graphs as new arriving nodes and links need to be processed timely~\cite{CAW}. However, the inductive learning capability on temporal graphs is less studied as opposed to that on static graphs.
Overall, the above challenges limit the applications of current research on large temporal graphs and remain to be addressed.

In light of these challenges, we propose SpikeNet, an inductive framework for modeling temporal graphs with high scalability. SpikeNet incorporates the spiking neural networks (SNNs) \cite{GerstnerK02} into the design of GNNs, with the aim to make better use of the spatio-temporal information while addressing the scalability issue on large temporal graphs. The key idea behind SpikeNet is the use of SNNs, a class of brain-inspired computing models that communicate with discrete spikes in an event-based manner. At each time step, SpikeNet aggregates and updates the signals (i.e., spikes) from a sampled neighborhood via an integrate-and-fire event. In this way, SpikeNet benefits from the spatio-temporal characteristics of SNNs with low computation and memory overheads. In summary, our contributions are as follows:
\begin{itemize}
    \item SpikeNet, a spike-based inductive learning framework for efficiently capturing the structural patterns and evolving dynamics in temporal graphs.
    \item A temporal architectural design based on SNNs, combining analog computation with spike-based propagation for more efficient model training and inference. To our best knowledge, this is the first work to build a general SNNs-based framework for modeling temporal graphs.
    \item An adaptive threshold update strategy for SNNs training to fit the neuron dynamics, which increases the flexibility and expressiveness of the resulting model.
    \item Extensive experiments that demonstrate the outstanding performance over advanced methods in real-world datasets on the temporal node classification tasks.
\end{itemize}

\section{Related Work}
In line with the focus of our work, we provide a brief overview of the background and related work on dynamic graph representation learning and spiking neural networks.

\subsection{Dynamic Graph Representation Learning}

Over the past few years, learning dynamic representations for temporal graphs has attracted considerable research effort \cite{dynamic_graph_survey}. Typically, methods developed for temporal graphs are often extensions of those for static ones, which additionally consider the temporal dimension and update schemes~\cite{EvolveGCN,tgat}.

Owning to the success of RNNs~\cite{cho2014learning}, one way to effectively model the temporal graph data is by employing the RNN (typically LSTM~\cite{hochreiter1997long}) architectures. Methods whereby RNNs generalize traditional GNNs to various temporal tasks in different manners~\cite{JODIE,EvolveGCN,GAEN,DBLP:journals/corr/abs-2006-10637, DBLP:conf/sigir/0001GRTY20,STAR}. For example, TGN~\cite{DBLP:journals/corr/abs-2006-10637} and JODIE~\cite{JODIE} update the memory and hidden state of the nodes by utilizing the LSTM and RNN units, and then the interaction events (node-wise or edge-wise) can be reasonably formulated.
EvolveGCN~\cite{EvolveGCN} proposes to use RNNs to regulate the GNN model parameters at every time step. Although the RNN-based methodology can efficiently save the history of interaction events related to the time dimension and capture graph dynamics, high dimensional vectors and numerous memory units also make the models suffer from massive computational costs and memory footprints.

Another line of research to explore the temporal graph dynamics is modeling sequential asynchronous discrete events occurring in continuous time as stochastic processes~\cite{DyRep,HTNE,MMDNE,DynamicTriad}. These methods try to capture evolution patterns over historical structures by a temporal point process, in addition to attention mechanisms~\cite{HTNE,MMDNE,DyRep} and temporal smoothness~\cite{DynamicTriad} enforced on the representation of continuous snapshots. However, the temporal point process may not allow the sharp change with regard to the node insertion and deletion (evolutionary), and the estimation models are typically quadratic in the number of interaction events, which is too computationally expensive for large graphs.

\subsection{Spiking Neural Networks}
SNNs are a class of brain-inspired and energy-efficient networks, which have attached great importance due to the distinctive properties of low power consumption and biological plausibility~\cite{DBLP:conf/aaai/BuDY022}. Different from artificial neural networks (ANNs) that make use of float values, SNNs deliver binary and asynchronous information through spikes only when the membrane potential reaches the threshold~\cite{kim2020spiking,DBLP:conf/aaai/BuDY022}. In literature, SNNs have been reported to offer huge energy-efficiency advantage over ANNs while achieving comparative results in a wide range of vision tasks, such as image classification~\cite{PLIF}, object detection~\cite{kim2020spiking}, and semantic segmentation~\cite{kim2021beyond}

Recently, efforts have been made towards applying SNN to graph domain~\cite{DBLP:conf/ijcnn/DoldG21,DBLP:journals/corr/abs-2109-10376,zhu2022spiking,ijcai2021-441}. Similar to that in vision research, they transform initial node features into a series spike trains via Poisson rate coding, followed by a graph convolution layer with SNN neurons. SNNs are inherently designed for temporal data, however, prior work mainly focuses on static graphs while ignoring the dynamic nature of SNNs. In this regard, the feasibility and advantages of SNNs for representing temporal graphs remain largely unexplored. This paper presents the first research effort to exploit SNNs on temporal graphs, which opens up possibilities for exploiting graph dynamics while avoiding high overheads.

\section{Preliminary}
In this section, we begin by giving the problem definition of this work, followed by the introduction of the leaky integrate-and-fire model.

\subsection{Problem Definition}\label{sec:def}
\paragraph{Notations}
In this work, we consider the discrete-time scenario of dynamic graph representation learning.
A temporal graph, in general, is defined as a sequence of graph snapshots at $T$ different time steps, denoted by $\mathcal{G} = \{\mathcal{G}^1, \mathcal{G}^2, \ldots, \mathcal{G}^T\}$\footnote{In this paper, we use the superscript $t$ to denote the time index.}. Each graph snapshot at time $t$ is defined as $\mathcal{G}^t=(\mathcal{V}^t,\mathcal{E}^t)$ where $\mathcal{V}^t=\{v_1,v_2,\ldots,v_N\}$ is a set of $N$ nodes and $\mathcal{E}^t\subseteq \mathcal{V}^t\times \mathcal{V}^t$ is a set of edges. Let $\mathcal{V}$ be the set of all nodes that appear in $\mathcal{G}$. Without loss of generality, we assume that the graph snapshot at any time is built on a common node set $\mathcal{V}$, in which the nonexistent node is treated as a dangling one with zero degree. In each time step, nodes can be paired with evolving features $X^t=\{x_v\ |\ \forall{v} \in \mathcal{V}\} \in \mathbb{R}^{N \times d}$, where $d$ is the feature dimensionality and is commonly stable over time. The entire node features with time dimension can be denoted as $\mathcal{X}=\{X^1,X^2,\ldots,X^T\} \in \mathbb{R}^{T \times N \times d}$.

Our goal is to learn node embedding $Z$ for all nodes that appeared in a temporal graph $\mathcal{G}$. The embedding should include nodes' (edges') evolution dynamics over time, which can be further used for downstream tasks like temporal node classification.

\subsection{Leaky Integrate-and-Fire Model}
In literature, the integrate-and-fire (IF)~\cite{IF} model is a mainstream computational model for neuron simulation. The IF model represents the membrane potential as a charge stored on a capacitor and abstracts the neuron dynamics of SNNs as three temporal events: (i) \textit{Integrate}. The neuron integrates current by means of the capacitor over time, which leads to a charge accumulation; (ii) \textit{Fire}. When the membrane potential has reached or exceeded a given threshold $V_\text{th}$, it fires (i.e., emits a spike). (iii) \textit{Reset}. Once fired, the membrane potential will be reset back to a constant value $V_\text{reset} < V_\text{th}$ like a biological neuron~\cite{Izhikevich04}.

However, neuron membranes are not perfect capacitors, rather they slowly leak current over time, pulling the membrane voltage back to its resting potential~\cite{Hunsberger}. In this regard, the LIF model additionally introduces a leaky term and describes the neuronal activity as an integration of received spike voltages as well as weak dissipation to the environment:
\begin{equation}
    \label{eq:LIF_diff}
    \tau_m \frac{d V}{d t} = -(V - V_\text{reset}) + \Delta V_m,
\end{equation}
where $\Delta V_m$ is the pre-synaptic input to the membrane voltage; $\tau_m$ is a membrane-related hyperparameter to control how fast the membrane potential decays, which leads to the membrane potential charges and discharges exponentially in response to the inputs. To better describe the neuron behaviors and guarantee computational availability, the differential equation in Eq.~\eqref{eq:LIF_diff} can also be converted to an iterative expression, as in~\cite{PLIF,zhu2022spiking}:
\begin{equation}
    \label{eq:LIF}
    V^{t} = V^{t-1} + \frac{1}{\tau_m}\left(-(V^{t-1} - V_\text{reset}) + {I}^t\right),
\end{equation}
where ${I}^t$ represents the pre-synaptic input from preceding neurons at time step $t$.

\section{Present Work: SpikeNet}

\begin{figure}[t]
    \centering
    \includegraphics[width=\linewidth]{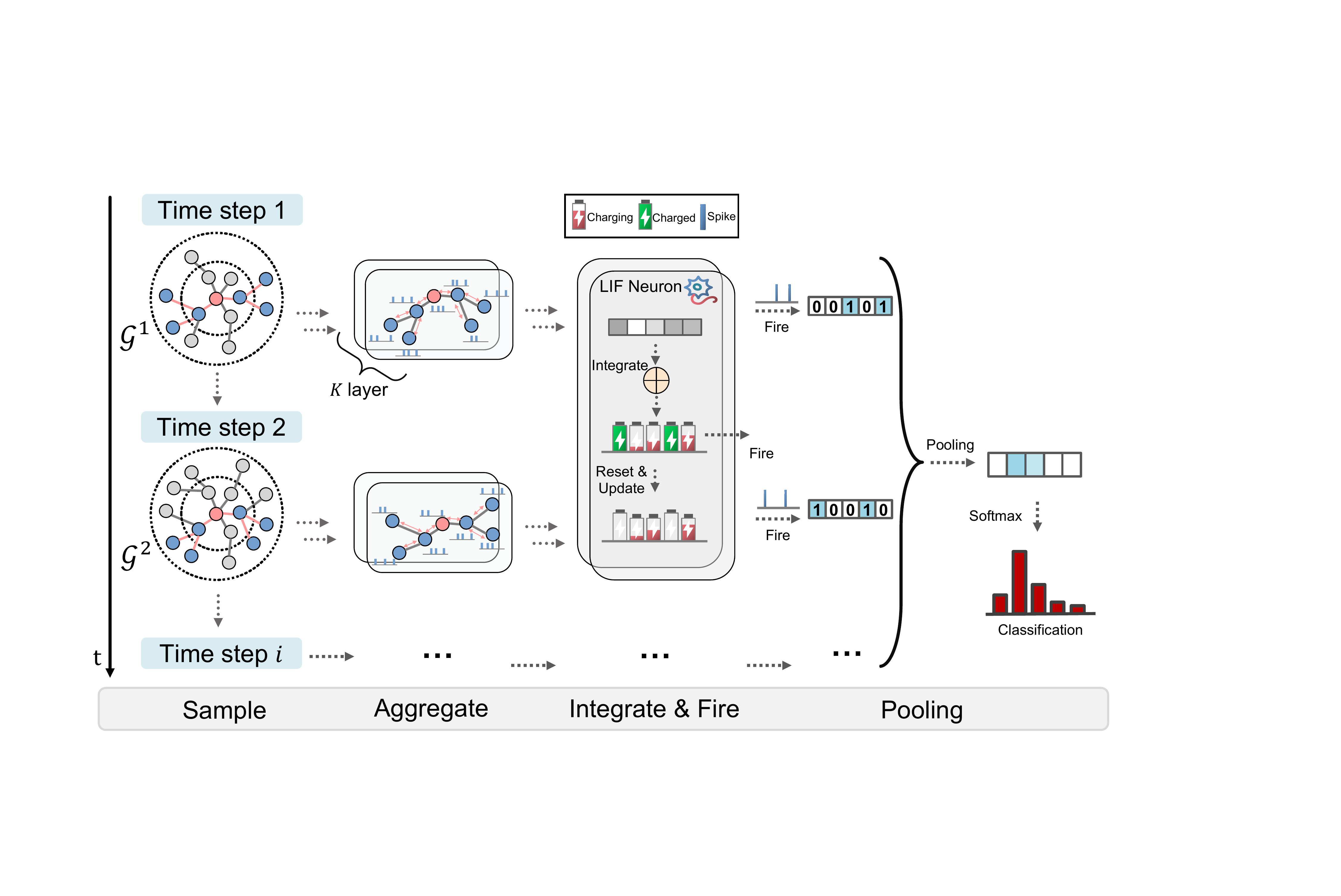}
    \caption{An overview of the SpikeNet framework.}
    \label{fig:framework}
\end{figure}

In this section, we propose SpikeNet, a scalable framework that generalizes spiking neural networks
to temporal graphs. SpikeNet is able to capture the evolving dynamics in temporal graphs, in which a variety of (unseen) nodes and edges are appeared/disappeared over time. An overview of SpikeNet is shown in Figure~\ref{fig:framework}. For each arriving timestamp, SpikeNet samples a subset of nodes and learns to aggregate spike signals from a node's local neighborhood. Then, the subsequent LIF model takes the aggregated signals as inputs to capture the temporal dynamics through an integrate-and-fire mechanism. Finally, the node embedding is obtained by taking historical spikes in each time step with a spike pooling operation, which can be used for downstream learning in a supervised manner.

\subsection{LIF as Temporal Architecture}
\label{sec:generalized_snn}
We explore a new direction in that we can use SNNs (i.e., LIF) rather than RNNs as the temporal architecture to capture the dynamic patterns in a graph sequence. Recall that the LIF's neuron behaviors are characterized by a series of events: \textit{integrate}, \textit{fire}, and \textit{reset}. For an arriving timestamp, each neuron in the LIF model updates the membrane potential based on its memorized state and current inputs, and then fires a spike when the membrane potential reaches a threshold $V_\text{th}$. If a spike is fired, the membrane potential is reset to a specific level $V_\text{reset}$, and the process starts again for the next round. Neurons at each layer undergo this process based on the input signals received from the preceding layer. Accordingly, a spike train is produced by each neuron for the subsequent layer.

However, prior work~\cite{PLIF,zhu2022spiking} typically adopts a fixed firing threshold $V_\text{th}$ for each LIF neuron across different layers, potentially limiting its flexibility and expressiveness. To address this issue, we propose an adaptive update strategy for neuron threshold, which calculates changes in neuron dynamics based on the previous threshold and incoming spikes:
\begin{equation}
    V^{t}_\text{th} = \tau_\text{th} V^{t-1}_\text{th} + \gamma O^{t},
\end{equation}
where $\tau_\text{th}$ and $\gamma$ are the decay factors to tune the threshold during training. $O^t$ is the spike train that takes a binary value (0 or 1) representing whether a neuron is spiked at time $t$. In this way, the firing threshold can be adaptively adjusted according to the neuron dynamics. We observe that the adaptive strategy also benefits the final performance in practice. By incorporating the threshold-triggered firing mechanism, membrane reset, and adaptive threshold update rules, we propose to generalize the computation scheme of the LIF model as follows:
\begin{align}
    \textbf{Integrate:} & \quad \quad V^{t}  = f(V^{t-1},{I}^t),                                          \\
    \textbf{Fire:}      & \quad \quad O^{t}  = \Theta({V^{t} - V^{t-1}_\text{th}}),                       \\
    \textbf{Reset:}     & \quad \quad V^{t}  = O^{t} V_\text{reset} + (1-O^{t})V^{t},                     \\
    \textbf{Update:}    & \quad \quad V^{t}_\text{th}  = \tau_\text{th} V^{t-1}_\text{th} + \gamma O^{t},
\end{align}
where $f$ denotes the integration behaviors of the LIF model when the neuron receives synaptic inputs from previous neurons, as in Eq.~\eqref{eq:LIF}. The decision to fire and generate a spike in the neuron output is carried out according to the Heaviside step function~\cite{PLIF}, which is defined by $\Theta(x)= 1$ if $x \geq 0$ and $0$ otherwise. For convenience, we simplify the above procedure as $O^{t}=\delta({I}^t)$ with $\delta(\cdot)$ the LIF model when receiving an input ${I}^t$. For each time step $t$, the input $I^t$ is the aggregated neighborhood information while the output $O^t$ is a spike train for intermediate node representations. In what follows, we will introduce how SpikeNet aggregates neighborhood messages as ${I}^t$ at each layer and time step $t$.

\subsection{Temporal Neighborhood Sampling}
\label{sec:sampler}
Neighborhood sampling is very important to learn node representations for large-scale graphs~\cite{HamiltonYL17}. Typically, it follows the design philosophy of directly sampling a set of nodes from multiple hops instead of using full neighborhood sets to avoid the over-expansion issue.
However, it was initially designed for static graphs without considering the structural evolution of temporal graphs.

In this work, we propose temporal neighborhood sampling on graph snapshots, which captures the graph dynamics from both macroscopic and microscopic perspectives. Given a node $v \in \mathcal{V}$, we sample a set of nodes by expanding the $v$'s neighborhood in the following way:
\begin{equation}
    \mathcal{S}^t(v) = \underbrace{\operatorname{SAMPLE}(v, \mathcal{G}^t)}_{\text{Macro-dynamics}} \cup \underbrace{\operatorname{SAMPLE}(v, \Delta \mathcal{G}^t)}_{\text{Micro-dynamics}},
\end{equation}
where $\operatorname{SAMPLE}$ is a graph sampler that produces the required neighborhood sets by uniform sampling \cite{HamiltonYL17} or random walk \cite{deepwalk}, to alleviate receptive field expansion and improve the computation efficiency. We denote $\mathcal{S}^t(v)$ a random sample of the node $v$'s neighbors at time step $t$. $\Delta \mathcal{G}^{t}=\mathcal{G}^{t} - \mathcal{G}^{t-1}$ represents a graph snapshot with all edges established at time $t$; Particularly $\Delta \mathcal{G}^{1}=\mathcal{G}^{1}$. The sampling step incorporates the \textit{macro-dynamics} and \textit{micro-dynamics} at each time step to learn node representations. Intuitively, micro-dynamics capture the fine-grained structural and temporal properties for node representation, while macro-dynamics explore the inherent evolution pattern of the graph from a global perspective.

\subsection{Spike-based Neighborhood Aggregation}
\label{sec:aggregator}
In the sampling step, we recursively sample the neighbors of root nodes up to a depth to perform message aggregation in a graph. At layer $k$, each node $v$ is related to a sequence of local neighborhood $\{\mathcal{S}^1,\mathcal{S}^2,\ldots,\mathcal{S}^{T}\}$ with $T$ time spans, to form a group of sample nodes along the time dimension. For each sampled set, SpikeNet computes the root node's hidden representation by recursively aggregating hidden node representations from bottom to top:
\begin{equation}
    \begin{aligned}
        h_{\mathcal{S}}^{t,(k)} & =\{h_{u}^{t,(k-1)}  \mathbf{W}, \forall u \in \mathcal{S}^t\},                                                                   \\
        h_{v}^{t,(k)}           & = \operatorname{\delta}\left(\operatorname{AGG}\left(\{h_{v}^{t,(k-1)} \mathbf{W}\} \cup h_{\mathcal{S}}^{t,(k)} \right)\right),
    \end{aligned}
\end{equation}
where $h_{v}^{t,(k)}$ is the hidden representation of node $v$ in $k$-th layer at time step $t$ and particularly $h_{v}^{t,(0)}=x_{v}^t$; $\mathbf{W}$ is the learnable parameter. $\operatorname{AGG}$ is a differentiable aggregation function, which aggregates features of local neighborhoods and passes them to the target node $v$. The aggregation function should be invariant to the permutations of node orderings such as a mean, sum, or max function. $\operatorname{\delta}(\cdot)$ is the LIF model which takes the aggregated information as input and outputs a spike train according to the neuron dynamics.

\begin{figure}[t]
    \centering
    \includegraphics[width=\linewidth]{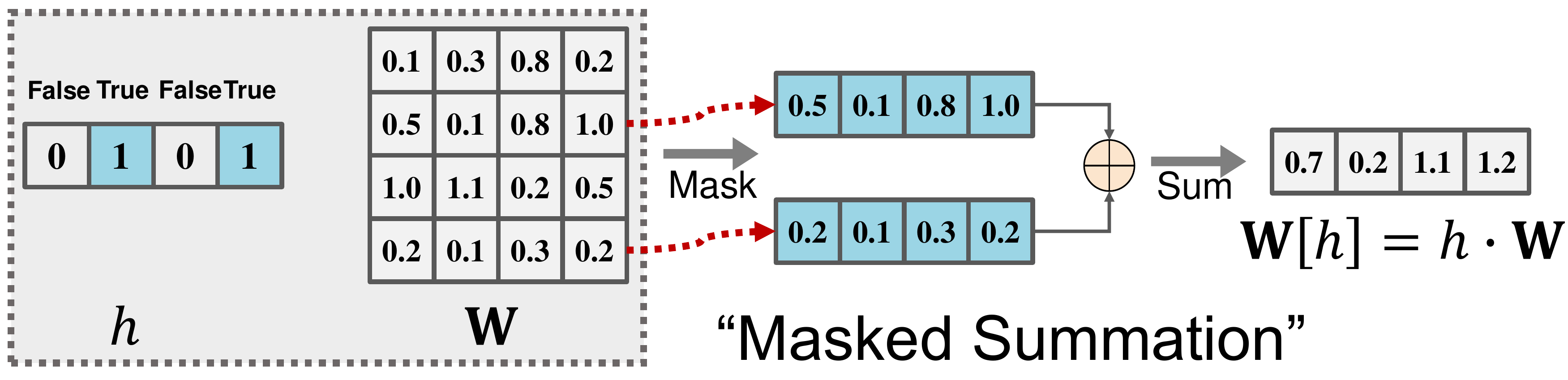}
    \caption{An example of ``Masked Summation'' operation $\mathbf{W}[h]$, which acts like matrix multiplication $h\cdot \mathbf{W}$ but enables more efficient computation.}
    \label{fig:masked_summation}
\end{figure}

In principle, the node representations are mapped to spike trains, offering a natural transition from the graph domain to SNNs. More specifically, $h_v$ is the pre-synaptic input denoted by a set of boolean numbers (spike or no spike), thus the matrix multiplication could be alternatively replaced by simple masking (indexing) operation combined with accumulation (since spikes are zero or one, a spike times any value $x$ is either zero or $x$, avoiding the need for explicit matrix multiplication):
\begin{equation}
    \begin{aligned}
        h_{\mathcal{S}}^{t,(k)} & =\{\mathbf{W}[h_{u}^{t,(k-1)}], \forall u \in \mathcal{S}^t\},                                                                    \\
        h_{v}^{t,(k)}           & = \operatorname{\delta}\left(\operatorname{AGG}\left(\{\mathbf{W}[h_{v}^{t,(k-1)}]\} \cup h_{\mathcal{S}}^{t,(k)} \right)\right),
    \end{aligned}
\end{equation}
where $\mathbf{W}[\cdot]$ denotes the \textit{Masked Summation} operation on $\mathbf{W}$. As shown in Figure~\ref{fig:masked_summation}, the row in a matrix $\mathbf{W}_{i,:}$ is masked (indexed) if the corresponding element $h_{i}=1(\text{True})$, and then the indexed row vectors are accumulated to calculate the summations. The operation can avoid expensive matrix multiplication computations entirely, and theoretically and also in practice, speed up the model.

\subsection{Spike Pooling}
The pooling layer is widely used to compress a set of nodes into a compact representation~\cite{WuC0L22}.
Inspired by the graph pooling in GNNs, we propose spike pooling as an independent operation to produce coarsened node representations over historical spikes. In this way, the final node representation is obtained by taking historical spikes at the last layer $K$ with a pooling operation:
\begin{equation}
    \label{eq:pooling}
    z_v= \operatorname{Pool}\left(\{h_v^{t,(K)}, \ldots, h_v^{T,(K)}\}\right),\  \forall v \in \mathcal{V}
\end{equation}
where $K$ is the number of layers and $T$ is the number of time steps. There are several different ways to implement the $\operatorname{Pool}$ operator, such as $\operatorname{Sum}$ and $\operatorname{Avg}$ which take the sum or average sum over all historical spikes. However, either $\operatorname{Sum}$ or $\operatorname{Avg}$ typically treats historical spikes equally in a time window, which might hinder the model's fitting capability in temporal tasks. In this regard, we adopt a learnable linear transformation over historical spikes:
\begin{equation}
    z_v= \operatorname{Linear}\left(\{h_v^{t,(K)} \|\ldots \| h_v^{T,(K)}\}\right),\  \forall v \in \mathcal{V}
\end{equation}
where $\|$ is a concatenating operator. The linear transformation could be implemented with a single-layer feed-forward neural network, where masked summation is also available to offer better efficiency during training and inference.

\subsection{Surrogate Gradient based Learning}
Given the node embeddings $z_v$, $\forall v \in \mathcal{V}$, we apply a softmax activation to transform node embeddings into prediction results for downstream tasks, followed by a task-specific objective (e.g., cross-entropy loss) to learn useful, predictive representations.

\begin{figure}[t]
    \centering
    \includegraphics[width=\linewidth]{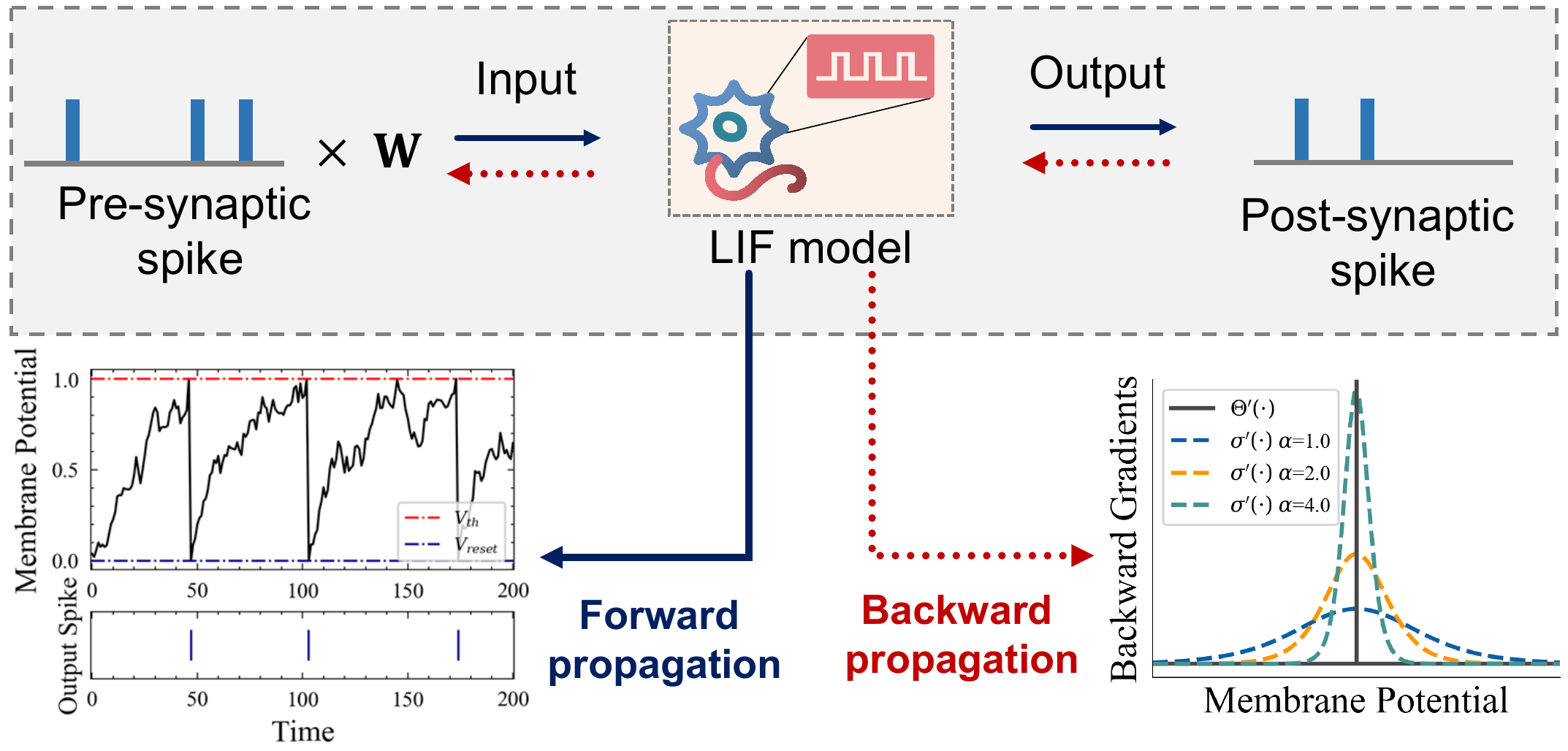}
    \caption{The illustration of forward propagation (blue arrow) and backward propagation (red dotted arrow) of SpikeNet. \textbf{Forward propagation}: the neuron outputs a spike via an integrate-and-fire mechanism, which brings the non-differentiability of the membrane potential. \textbf{Backward propagation}: a smooth Sigmoid function with different smooth factor $\alpha$ is adopted to approximate the backward gradients.}
    \label{fig:surrogate}
\end{figure}

Note that it is difficult to train SpikeNet directly via standard gradient descent due to the non-differentiability of discrete spikes and the hard threshold function. Inspired by the surrogate learning technique~\cite{DBLP:journals/corr/abs-1901-09948,PLIF}, we circumvent the non-differentiable backpropagation problem by defining a surrogate function for LIF neurons when calculating backward gradients. Here we use a smooth Sigmoid function~\cite{smooth_sigmoid_2} $\sigma(\alpha x)=1/(1+\text{exp}(-\alpha x))$ to calculate surrogate gradients during backward propagation, whose derivative is formally defined as:
\begin{equation}
    \begin{aligned}
        \sigma(\alpha x)^\prime & =\alpha \cdot \sigma(\alpha x) \cdot (1-\sigma(\alpha x)),
    \end{aligned}
\end{equation}
where $\alpha$ is a constant factor that controls the smoothness. In general, a larger $\alpha$ leads to a better approximation of the hard threshold function $\Theta(\cdot)$, but causes the vanishing and exploding gradient problem, which in turn makes the training very difficult to converge. Therefore, choosing a suitable value of the smooth factor $\alpha$ is critical for our algorithm to achieve comparable performance.

Figure~\ref{fig:surrogate} illustrates the surrogate learning procedure of SpikeNet. During the forward propagation, the membrane potential increases with pre-synaptic inputs, which are obtained from pre-synaptic spikes multiplied by the corresponding weight $W$. When the membrane potential reaches the threshold $V_\text{th}$, the neuron generates the post-synaptic spike, and then the membrane potential discharges to a resting potential $V_\text{reset}$. During the backward propagation, a smooth Sigmoid function with different $\alpha$ is adopted to implement the backward gradient to take advantage of the standard backpropagation-based optimization procedures.

\begin{table}[t]
    \centering
    \begin{tabular}{l|ccc}
        \toprule
                     & \textbf{DBLP} & \textbf{Tmall} & \textbf{Patent} \\
        \midrule
        \#Nodes      & 28,085        & 577,314        & 2,738,012       \\
        \#Edges      & 236,894       & 4,807,545      & 13,960,811      \\
        \#Classes    & 10            & 5              & 6               \\
        \#Time steps & 27            & 186            & 25              \\
        \bottomrule
    \end{tabular}
    \caption{Dataset Statistics.}\label{tab:data}
\end{table}

\begin{table*}[t]
    \centering
    \resizebox{\textwidth}{!}{\begin{tabular}{c|c|c|ccccccccc|c}
            \toprule
            {\textbf{Dataset}}               & {\textbf{Metrics}}          & \textbf{Tr.ratio} & \textbf{DeepWalk} & \textbf{Node2Vec} & \textbf{HTNE} & \textbf{M$^2$DNE} & \textbf{DyTriad} & \textbf{MPNN}   & \textbf{JODIE}  & \textbf{EvolveGCN}       & \textbf{TGAT}   & \textbf{SpikeNet}        \\
            \midrule
            \multirow{6}{*}{\textbf{DBLP}}   &
            \multirow{3}{*}{{Macro-F1}}      & 40\%                        & 67.08             & 66.07             & 67.68             & 69.02         & 60.45             & 64.19${\pm0.4}$  & 66.73${\pm1.0}$ & 67.22${\pm0.3}$ & \textbf{71.18${\pm0.4}$} & 70.88${\pm0.4}$                            \\
                                             &                             & 60\%              & 67.17             & 66.81             & 68.24         & 69.48             & 64.77            & 63.91${\pm0.3}$ & 67.32${\pm1.1}$ & 69.78${\pm0.8}$          & 71.74${\pm0.5}$ & \textbf{71.98${\pm0.3}$} \\
                                             &                             & 80\%              & 67.12             & 66.93             & 68.36         & 69.75             & 66.42            & 65.05${\pm0.5}$ & 67.53${\pm1.3}$ & 71.20${\pm0.7}$          & 72.15${\pm0.3}$ & \textbf{74.65${\pm0.5}$} \\
            \cmidrule{2-13}
                                             & \multirow{3}{*}{{Micro-F1}} & 40\%              & 66.53             & 66.80             & 68.53         & 69.23             & 65.13            & 65.72${\pm0.4}$ & 68.44${\pm0.6}$ & 69.12${\pm0.8}$          & 71.10${\pm0.2}$ & \textbf{71.98${\pm0.5}$} \\
                                             &                             & 60\%              & 66.89             & 67.37             & 68.57         & 69.47             & 66.80            & 66.79${\pm0.6}$ & 68.51${\pm0.8}$ & 70.43${\pm0.6}$          & 71.85${\pm0.4}$ & \textbf{72.35${\pm0.8}$} \\
                                             &                             & 80\%              & 66.38             & 67.31             & 68.79         & 69.71             & 66.95            & 67.74${\pm0.3}$ & 68.80${\pm0.9}$ & 71.32${\pm0.5}$          & 73.12${\pm0.3}$ & \textbf{74.86${\pm0.5}$} \\
            \midrule
            \multirow{6}{*}{\textbf{Tmall}}  & \multirow{3}{*}{{Macro-F1}} & 40\%              & 49.09             & 54.37             & 54.81         & 57.75             & 44.98            & 47.71${\pm0.8}$ & 52.62${\pm0.8}$ & 53.02${\pm0.7}$          & 56.90${\pm0.6}$ & \textbf{58.84${\pm0.4}$} \\
                                             &                             & 60\%              & 49.29             & 54.55             & 54.89         & 57.99             & 48.97            & 47.78${\pm0.7}$ & 54.02${\pm0.6}$ & 54.99${\pm0.7}$          & 57.61${\pm0.7}$ & \textbf{61.13${\pm0.8}$} \\
                                             &                             & 80\%              & 49.53             & 54.58             & 54.93         & 58.47             & 51.16            & 50.27${\pm0.5}$ & 54.17${\pm0.2}$ & 55.78${\pm0.6}$          & 58.01${\pm0.7}$ & \textbf{62.40${\pm0.6}$} \\
            \cmidrule{2-13}
                                             & \multirow{3}{*}{{Micro-F1}} & 40\%              & 57.11             & 60.41             & 62.53         & \textbf{64.21}    & 53.24            & 57.82${\pm0.7}$ & 58.36${\pm0.5}$ & 59.96${\pm0.7}$          & 62.05${\pm0.5}$ & 63.52${\pm0.7}$          \\
                                             &                             & 60\%              & 57.34             & 60.56             & 62.59         & 64.38             & 56.88            & 57.66${\pm0.5}$ & 60.28${\pm0.3}$ & 61.19${\pm0.6}$          & 62.92${\pm0.4}$ & \textbf{64.84${\pm0.4}$} \\
                                             &                             & 80\%              & 57.88             & 60.66             & 62.64         & 64.65             & 60.72            & 58.07${\pm0.6}$ & 60.49${\pm0.3}$ & 61.77${\pm0.6}$          & 63.32${\pm0.7}$ & \textbf{66.10${\pm0.3}$} \\
            \midrule
            \multirow{6}{*}{\textbf{Patent}} & \multirow{3}{*}{{Macro-F1}} & 40\%              & 72.32${\pm0.9}$   & 69.01${\pm0.9}$   & -             & -                 & -                & -               & 77.57${\pm0.8}$ & 79.67${\pm0.4}$          & 81.51${\pm0.4}$ & \textbf{83.53${\pm0.6}$} \\
                                             &                             & 60\%              & 72.25${\pm1.2}$   & 69.08${\pm0.9}$   & -             & -                 & -                & -               & 77.69${\pm0.6}$ & 79.76${\pm0.5}$          & 81.56${\pm0.6}$ & \textbf{83.85${\pm0.7}$} \\
                                             &                             & 80\%              & 72.05${\pm1.1}$   & 68.99${\pm1.0}$   & -             & -                 & -                & -               & 77.67${\pm0.4}$ & 80.13${\pm0.4}$          & 81.57${\pm0.5}$ & \textbf{83.90${\pm0.6}$} \\
            \cmidrule{2-13}
                                             & \multirow{3}{*}{{Micro-F1}} & 40\%              & 71.57${\pm1.3}$   & 68.14${\pm0.9}$   & -             & -                 & -                & -               & 77.64${\pm0.7}$ & 79.39${\pm0.5}$          & 80.79${\pm0.7}$ & \textbf{83.48${\pm0.8}$} \\
                                             &                             & 60\%              & 71.53${\pm1.0}$   & 68.20${\pm0.7}$   & -             & -                 & -                & -               & 77.89${\pm0.5}$ & 79.75${\pm0.3}$          & 80.81${\pm0.6}$ & \textbf{83.80${\pm0.7}$} \\
                                             &                             & 80\%              & 71.38${\pm1.2}$   & 68.10${\pm0.5}$   & -             & -                 & -                & -               & 77.93${\pm0.4}$ & 80.01${\pm0.3}$          & 80.93${\pm0.6}$ & \textbf{83.88${\pm0.9}$} \\
            \bottomrule
        \end{tabular}
    }
    \caption{Quantitative results (\%) on the temporal node classification task. The results are averaged over five runs, where the best results in each row are highlighted in \textbf{boldfaced}. (Tr.ratio: training ratio)} \label{table:nodeclas}

\end{table*}

\subsection{Time Complexity}
Here we briefly discuss the time complexity of our proposed SpikeNet. For simplicity, we assume that the node feature and hidden representations are $d$-dimensional, and the neighborhood size is fixed as $|\mathcal{S}|$ and shared by each hop. As the LIF model and spike pooling are both simple and computationally efficient, the main bottleneck is the aggregation step involved in all $T$ graph snapshots.
Particularly, for a $K$ layer SpikeNet, the time complexity is related to the sampled graph size and the dimension of features/hidden representations, about $\mathcal{O}(T|\mathcal{V}||\mathcal{S}|^Kd^2)$. Since $|\mathcal{S}|$, $K$, and $d$ are often small, the computation overhead is acceptable even for large graphs. In addition, the time complexity could be further reduced by implementing the masked summation operation in specialized chips.

\section{Experiments}
In this section, we conduct experiments on three large real-world graph datasets: DBLP, Tmall~\cite{MMDNE}, and Patent~\cite{patent}. The datasets statistics are listed in Table \ref{tab:data}. Limited by space, we present the details about the datasets and experimental settings in Appendix.

\subsection{Overall Performance}
In our experiments, we consider the task of temporal node classification to evaluate the representation quality. In this task, the graph structure at each snapshot is entirely available for representation learning. We follow \cite{MMDNE} and examine the performance when different sizes of training datasets are used, i.e., 40\%, 60\%, and 80\%, including 5\% for validation. For unsupervised methods, DeepWalk, Node2Vec, HTNE, M$^2$DNE, and DyTriad, a two-layer MLP with hidden units 128 are trained to solve the classification problem. Table \ref{table:nodeclas} summarizes the performance of all the compared methods on the temporal node classification task.  We repeat the experiment for five runs with different random seeds and report the average results with standard deviation in terms of both Macro-F1 and Micro-F1 scores. We reuse the results of DeepWalk, Node2Vec, HTNE, M$^2$DNE, and DyTriad on DBLP and Tmall already reported in \cite{MMDNE}.

As shown in Table \ref{table:nodeclas}, methods developed for static graphs (i.e., DeepWalk and Node2Vec) generally underperform those for temporal graphs, which suggests that the temporal dynamics of the graph structure are indeed critical to learn a better node representation. We can also observe that the performance of unsupervised methods remains stable as the ratio of the training set increases. By contrast, MPNN, TGAT, EvolveGCN, JODIE, and SpikeNet can benefit from the increasing training data. This is due to the fact that supervised methods can incorporate supervision signals into learning and facilitate training. In this regard, supervised methods are also capable of outperforming most unsupervised ones. On the largest graph dataset Patent, HTNE, M$^2$DNE, and DyTriad fail to learn properly since the estimation process is time-consuming and thus too expensive to model large graphs. MPNN, an LSTM-based approach is also not available for Patent, since it requires high computation overheads to train the model in a full-batch fashion.

Although SpikeNet uses discrete spiking signals to communicate between layers rather than real-valued, it achieves state-of-the-art performance in most cases. In particular, SpikeNet slightly underperforms TGAT and M$^2$DNE on DBLP and Tmall when the training data is small.
As the training data increases, SpikeNet is able to outperform strong baselines. Notably, SpikeNet achieves about 4.7\% and 4.3\% performance improvement on average over the strongest baseline in Tmall and Patent as measured by Micro-F1, respectively. One potential reason could be that the surrogate learning technique requires more training data to better approximate the backward gradients. In addition, more training data can also help the threshold update strategy to adjust the neuron thresholds adaptively. The results indicate that the SNNs-based framework provides a simple yet effective way for dynamic graph representation learning.

\subsection{Overhead Evaluation}
In this section, we compare the overheads between SpikeNet and other baseline approaches
that are most relevant to.

\subsubsection{Parameter size}
\begin{figure}[t]
    \centering
    \subfigure[Parameter size]{\includegraphics[width=0.48\linewidth,height=0.33\hsize]{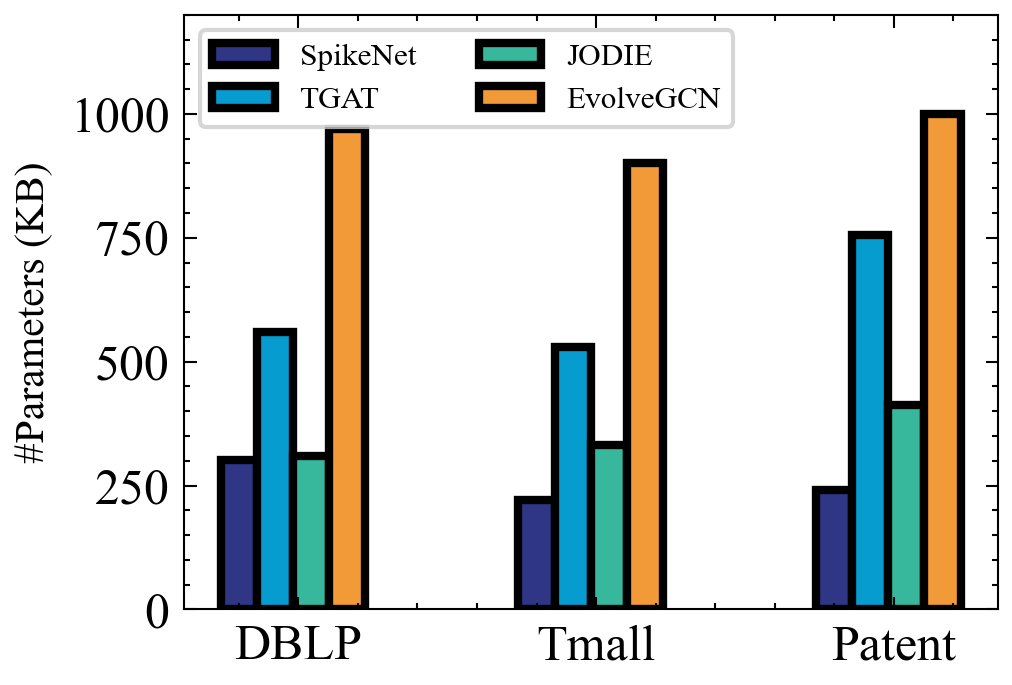}
    }
    \subfigure[Training time]{\includegraphics[width=0.48\linewidth,height=0.33\hsize]{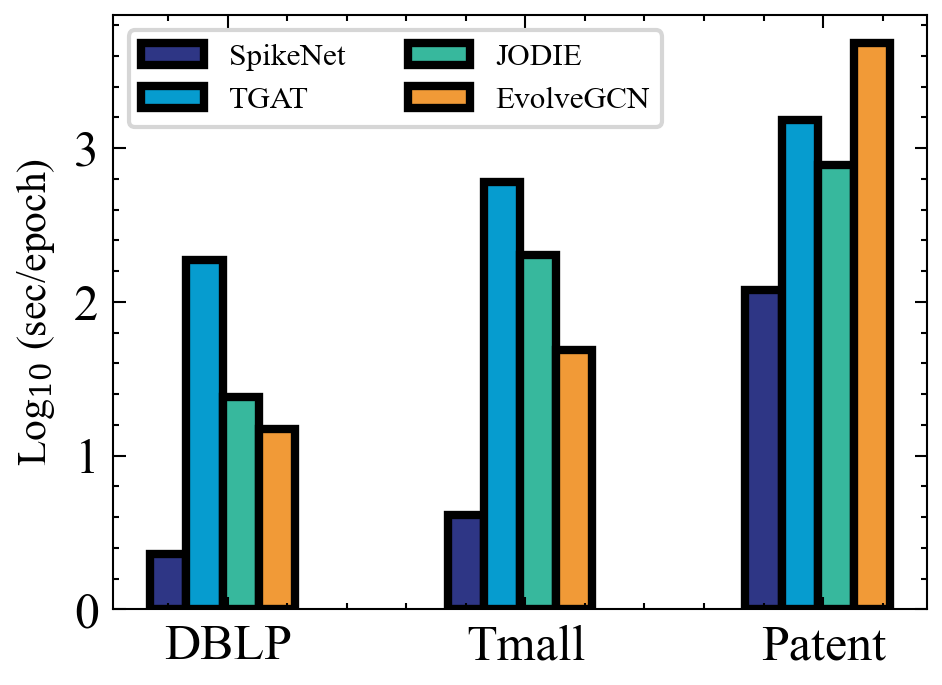}
    }
    \caption{Overheads of different methods in terms of parameter size and training time. Methods that cannot run on Patent are excluded for comparison.}
    \label{fig:complexity}

\end{figure}

The number of parameters of different methods is shown in Figure~\ref{fig:complexity}(a).
It can be seen that SpikeNet is much more lightweight with only around 50\% parameters as compared to TGAT, which uses a multi-head attention mechanism and requires a lot of parameters to learn.  Particularly, EvolveGCN which combines GNNs with RNNs has around four times the parameter size than SpikeNet. However, more parameters would result in high memory overheads and the overfitting issue. In this regard, SpikeNet replaces the RNN cell with a parameter-free LIF model, which significantly reduces the number of parameters to be optimized almost without losing accuracy.

\subsubsection{Runtime complexity}
The training time of each epoch for different methods is shown in Figure~\ref{fig:complexity}(b). The experiments were run on a Titan RTX GPU with the same batch size (except for EvolveGCN which is fully batch trained) for a fair comparison.
Through the experiment, we find that SpikeNet holds obvious efficiency superiority during training. Specifically, SpikeNet can obtain orders of magnitude speedup over baselines. We believe the significant improvement can be attributed to the following two reasons.: (i) Compared to EvolveGCN, SpikeNet trains the model on a small sampled subgraph instead of the whole graph during training, so the training time is significantly smaller. (ii) Compared to JODIE and TGAT, SpikeNet benefits from the lightweight temporal architecture (i.e., LIF model), with much fewer computation overheads and parameters to be optimized. The simplification also speeds up the training.
Note that, the masked summation operation between two matrices is currently not well supported in the deep learning framework, so we use matrix multiplication instead to facilitate the implementation. In the near future, SpikeNet could also benefit from neuromorphic chips and speed up the implementation.

\subsection{Ablation Study}
\label{sec:abla}
To further investigate the impact of different hyperparameters in SpikeNet, we conduct several experiments to analyze them from different perspectives.

\subsubsection{Threshold decay $\tau_\text{th}$ and $\gamma$}
\begin{figure}[t]
    \centering
    \subfigure[DBLP]{\includegraphics[width=0.48\linewidth,height=0.33\hsize]{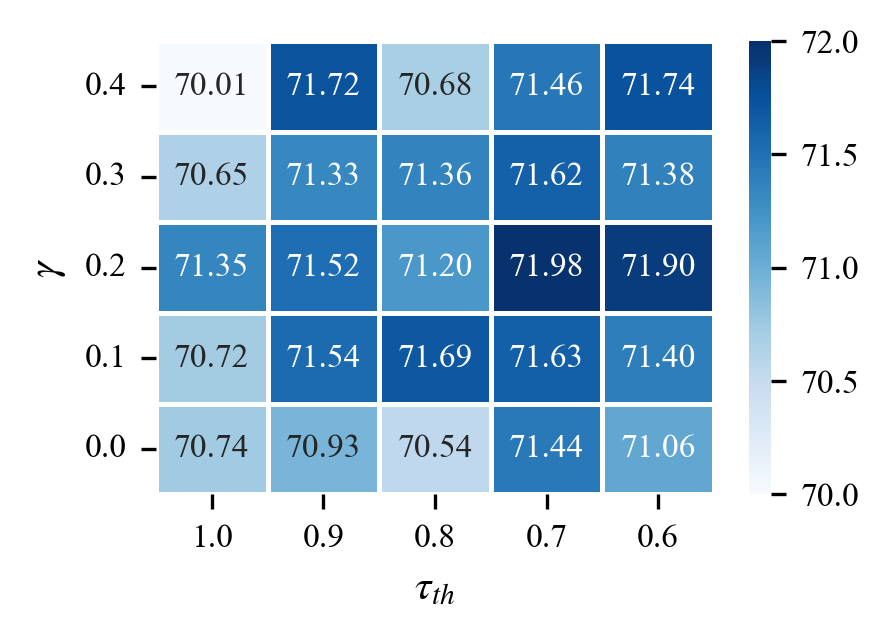}}
    \subfigure[Tmall]{\includegraphics[width=0.48\linewidth,height=0.33\hsize]{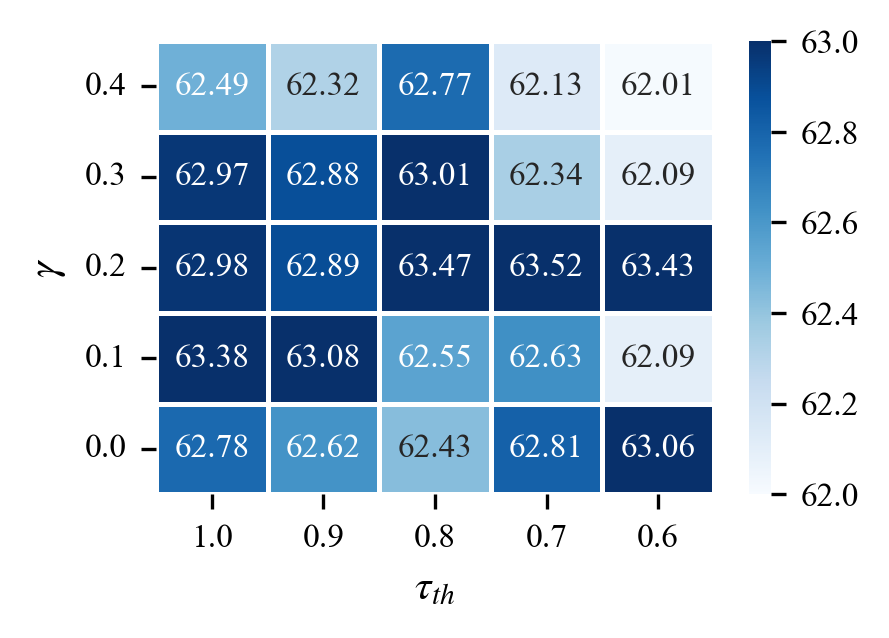}}
    \caption{Ablation analysis w.r.t. $\tau_\text{th}$ and $\gamma$ on DBLP and Tmall, respectively. Here we report Micro-F1 results with Tr.ratio=0.4 for both datasets. $\tau_\text{th}=1.0$ and $\gamma=0$ denote the standard case without thresholds decay.}
    \label{fig:threshold}
\end{figure}

As one of our main contributions, the threshold decay strategy is adopted for LIF model to stablize training. To comprehensively explore the effectiveness of threshold decay strategy, we cownduct temporal node classification on DBLP and Tmall by varying the values of $\tau_\text{th}$ and $\gamma$ as \{1.0, 0.9, 0.8, 0.7, 0.6\} and \{0., 0.1, 0.2, 0.3, 0.4\}, respectively.  The experiments are repeated 5 times and the average results are shown in Figure~\ref{fig:threshold}. It is observed that the threshold decay strategy generally benefit the learning of SpikeNet on two datasets, as evidenced by decreasing $\tau_\text{th}$ and increasing $\gamma$ simultaneously can improve the performance of SpikeNet. When $\tau_\text{th}=0.7$ and $\gamma=0.2$, SpikeNet achieves the best performance, which is much better than not using decay. Overall, the results demonstrate that the threshold decay strategy is beneficial for the model performance.

\subsubsection{Smooth factor $\alpha$}
\label{sec:alpha}
\begin{figure}[t]
    \centering
    \subfigure[$\Theta(\cdot)$ and $\delta(\cdot)$]{\includegraphics[width=0.48\linewidth,height=0.33\hsize]{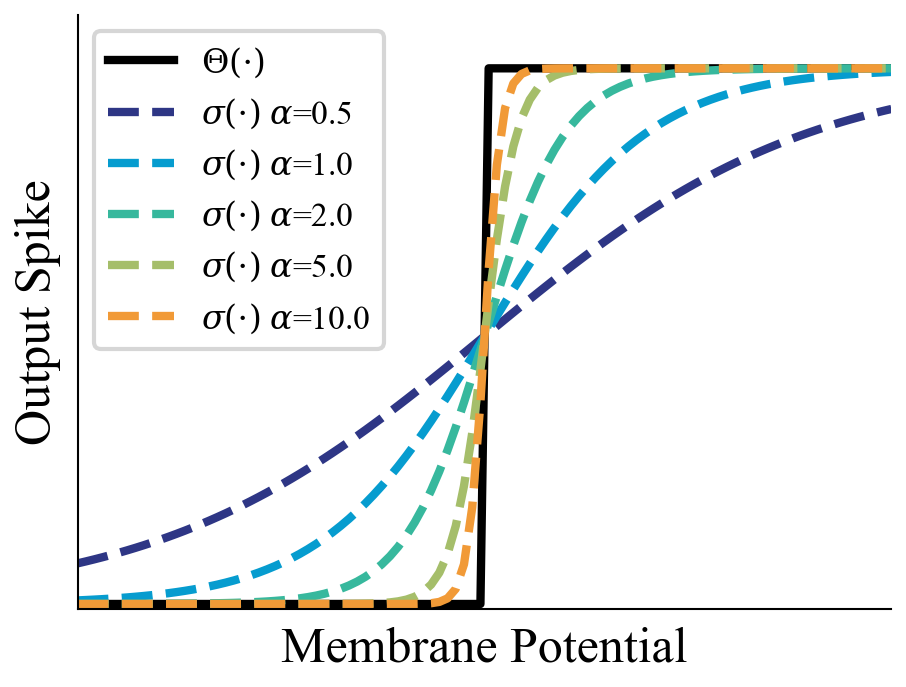}
    }
    \subfigure[Classification performance]{\includegraphics[width=0.48\linewidth,height=0.33\hsize]{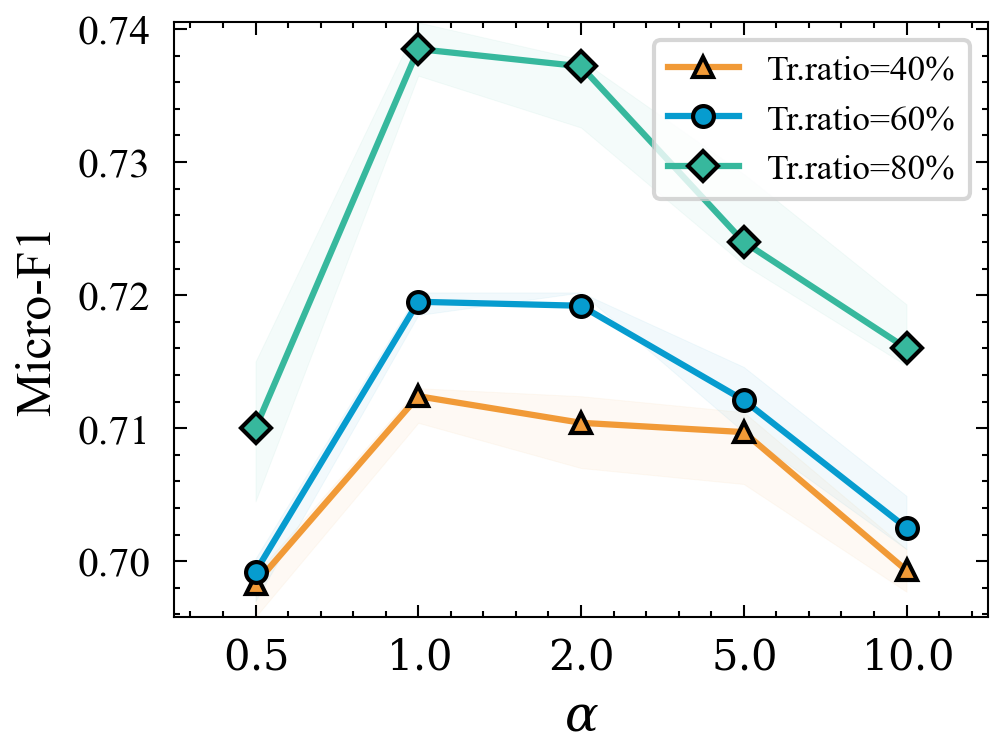}
    }
    \caption{Ablation analysis w.r.t. smooth factor $\alpha$ on DBLP.}
    \label{fig:smooth_factor}
\end{figure}

Since surrogate learning technique is important for training the SNN-based model, we explore the sensitivity of smooth factor $\alpha$ in SpikeNet. We vary the smooth factor $\alpha$ from \{0.5, 1.0, 2.0, 5.0, 10.0\} to study the effects in the surrogate function $\sigma(\cdot)$. We only report the results on DBLP, because we have similar observations for other datasets. As shown in Figure~\ref{fig:smooth_factor}(a), it is observed that $\sigma(\cdot)$ can better approximate the Heaviside step function $\Theta(\cdot)$ with the increase of $\alpha$. However, the best performance is achieved when $\alpha=1.0$ according to Figure~\ref{fig:smooth_factor}(b). This appears to be quite straightforward since higher $\alpha$ will suffer from the vanishing and exploding gradient problem. Overall, SpikeNet is sensitive to $\alpha$ and the choice of a good $\alpha$ is important to learn the model properly.

\section{Conclusion}
We present SpikeNet, a scalable framework built upon SNNs to capture the evolving dynamics underlying a discrete graph sequence. To our best knowledge, SpikeNet is the first research effort to exploit SNNs on temporal graphs. The key insight behind our approach is to aggregate and update temporal information from a dynamically sampled neighborhood via an integrate-and-fire mechanism. As a result, SpikeNet enjoys the advantage of exploiting graph dynamics as well as significantly fewer parameters and overheads.
SpikeNet inherently allows inductive learning, making it applicable to predict evolving dynamics on temporal graphs with unseen nodes. Experiments on three real-world datasets show that SpikeNet outperforms the recent state-of-the-art methods in most cases. Through further analysis of these results, we also show that SpikeNet provides better trade-offs between performance and computational costs particularly compared to RNN-based methods.



\section{Acknowledgments}
The research is supported by the Key-Area Research and Development Program of Guangdong Province (2020B010165003), the Guangdong Basic and Applied Basic Research Foundation (2020A1515010831), the Guangzhou Basic and Applied Basic Research Foundation (202102020881), and CCF-AFSG Research Fund (20210002). Qi Yu is supported in part by an NSF IIS award IIS-1814450 and an ONR award N00014-18-1-2875. The views and conclusions contained in this paper are those of the authors and should not be interpreted as representing any funding agency.

\bibliography{aaai23}

\clearpage
\appendix
\section{Algorithm}
\label{appendix:alg}

\begin{algorithm}[h]
    \caption{SpikeNet embedding generation algorithm (forward propagation)}
    \label{algorithm}
    \begin{algorithmic}[1] 
        \REQUIRE ~~\\
        Temporal graph $\mathcal{G} = \{\mathcal{G}^1, \mathcal{G}^2, \ldots, \mathcal{G}^T\}$; input node features $\mathcal{X}=\{X^1, X^2, \ldots, X^T\}$; depth $K$ and time span $T$; weight matrices $\mathbf{W}^{(k)}$, neighborhood sampler $\operatorname{SAMPLE}$, differentiable aggregation function $\operatorname{AGG}$ and spike pooling function $\operatorname{Linear}$; LIF models $\operatorname{\delta}^{(k)}$ $\forall k \in \{1,\ldots,K\}$;
        \ENSURE ~~\\
        Node representations $z_v$ for all $v \in \mathcal{V}$;

        \STATE $h_{v}^{t,(0)} \leftarrow x_v^{t}, \forall v \in \mathcal{V}, t\in \{1,\ldots,T\}$;
        \FOR {$t=1$ to $T$}{
        \FOR {$k=1$ to $K$}{
        \FOR {$v \in \mathcal{V}$}{

        \STATE \small $\mathcal{S}^t(v) \leftarrow {\operatorname{SAMPLE}(v, \mathcal{G}^t)} \cup {\operatorname{SAMPLE}(v, \Delta \mathcal{G}^t)}$;
        \STATE  $h_{\mathcal{S}}^{t,(k)} \leftarrow \operatorname{AGG}\left(\{\mathbf{W}^{(k)}[h_{u}^{t,(k-1)}], \forall u \in \mathcal{S}^t\}\right)$;
        \STATE $h_{v}^{t,(k)} \leftarrow \operatorname{\delta}^{(k)}\left(\mathbf{W}^{(k)}[h_{v}^{t,(k-1)}] + h_{\mathcal{S}}^{t,(k)}\right)$;
        }
        \ENDFOR
        }
        \ENDFOR
        }
        \ENDFOR
        \RETURN \small $z_v \leftarrow \operatorname{Linear}\left(\{h_v^{t,(K)} \|\ldots \| h_v^{T,(K)}\}\right),\  \forall v \in \mathcal{V}$;
    \end{algorithmic}
\end{algorithm}

Algorithm~\ref{algorithm} provides an overview of the SpikeNet Embedding generation (i.e., forward propagation), the node embedding at each layer and each time step is generated by aggregating neighborhood information, followed by a LIF model to capture the evolving dynamics. It is worth noting that the forward propagation allows the multiply-and-accumulate typically inherent in matrix multiplication to be turned into simply matrix masking (or indexing) and accumulation, i.e., masked summation, which could be implemented with more energy-efficient neuromorphic hardware such as Intel Loihi~\cite{davies2018loihi} or Brainchip Akida~\cite{vanarse2019hardware}. In this manner, our method may lead to more energy-efficient implementations of GNNs on temporal graphs and will be meaningful and influential in the future.

\section{Experimental setup}
\label{appendix:exp}

\subsection{Dataset Statistics}
We adopt three temporal graph datasets with different scales: DBLP, Tmall, and Patent.
We briefly give an overview of the datasets:

\textbf{DBLP}~\cite{MMDNE}. DBLP is an academic co-author graph extracted from the bibliography website. in which each node is an author and each edge means the two authors collaborated on a paper. The authors in DBLP are labeled according to their research areas.

\textbf{Tmall}~\cite{MMDNE}. Tmall is a bipartite graph extracted from the sales data in 2014 at Tmall.com, in which each node refers to either one user or one item and each edge refers to one purchase with a timestamp. The five most frequently purchased categories are treated as labels in experiments. Note that Tmall is a partially labeled dataset in which only items are assigned with categories. We follow the pre-processing scheme as described in~\cite{MMDNE} that selects the five most frequently purchased categories as labels in experiments.

\textbf{Patent}~\cite{patent}. Patent is a citation network of US patent ranging from year 1963 to 1999. Each patent belongs to six different types of patent categories. Patent is the largest size of graph among all three datasets.

\subsection{Baseline Methods}
We compare the performance of SpikeNet with the following baselines, including static and dynamic graph representation learning methods:

\begin{enumerate}
    \item \textbf{DeepWalk}~\cite{deepwalk} performs random walks to generate an ordered sequence of nodes from a static graph to create contexts for each node, then applies a skip-gram model to these sequences to learn representations.
    \item \textbf{Node2Vec}~\cite{node2vec} extends DeepWalk with biased random walks to explore the neighborhood of a node and learn node representations on a static graph.
    \item \textbf{HTNE}~\cite{HTNE} integrates the Hawkes process and attention mechanism into network embedding to model the neighborhood formation sequences.
    \item \textbf{M$^2$DNE}~\cite{MMDNE} designs a temporal attention point process and a general dynamics equation to capture structural and temporal properties of evolving graphs.
    \item \textbf{DynamicTriad} (DyTriad)~\cite{DynamicTriad} models both structural information and evolution patterns based on the triadic closure process, which enables the model to capture the graph dynamics effectively.
    \item \textbf{MPNN-LSTM} (MPNN)~\cite{DBLP:conf/aaai/PanagopoulosNV21} is a time-series version of message passing neural network (MPNN) with two-layer LSTM~\cite{hochreiter1997long}. MPNN can capture the long-range temporal dependencies in temporal graphs based on the dynamics encoded into the node representations.
    \item \textbf{JODIE}~\cite{JODIE} employs two RNNs to update the node embeddings at every observed interaction and uses a projection operation that predicts the future embedding trajectory in an evoloving graph.
    \item \textbf{EvolveGCN}~\cite{EvolveGCN} uses the RNN to evolve the GNN parameters at each time step, which effectively performs model adaptation and captures the dynamics of the graph sequence.
    \item \textbf{TGAT}~\cite{tgat} utilizes self-attention mechanisms as basic blocks and derive a functional time encoding to effectively learn temporal and topological information, which achieves state-of-the-art results for different dynamic graph learning tasks.
\end{enumerate}

For DeepWalk and Node2Vec that are originally designed for static graphs, we accumulate historical information in the graph sequence and represent the structure at the last time period, i.e., $\mathcal{G}^{T}$.  More importantly, there are several recent
works~\cite{GAEN,STAR,DBLP:conf/aaai/XuLCWCZ21,DBLP:journals/corr/abs-2006-10637,DBLP:conf/sigir/0001GRTY20} that are not considered for comparison in our experiments because they require very high memory footprints for modeling the temporal graphs and cannot run on a reasonably sized GPU (24GB).

\subsection{Implementation Details}
Each experiment is conducted five times and average results with standard deviation are reported. The hyperparameters are tuned based on the performance of validation set. More specifically, we implement DeepWalk and Node2Vec using the open-source library GraphGallery~\cite{li2021graphgallery} with an embedding size of 128. For dynamic methods, we use the code provided by the authors and closely follow the experimental setup in~\cite{MMDNE}, so as to reduce the experiment workload and make a fair comparison. For other parameters, we follow the setup in \cite{MMDNE}. 

For SpikeNet, it supports batch training for both transductive and inductive tasks where the batch size is tuned in [512, 1024, 2048, 4096]. The learning rate is ranged in [0.001, 0.0003, 0.005, 0.008, 0.01] with AdamW~\cite{AdamW} optimizer. We use two-layer structure for SpikeNet and the aggregate function $\operatorname{AGG}$ is set as \textit{mean}. We set the embedding size of 128, 512, and 512 for DBLP, Tmall, and Patent. For the temporal neighborhood sampling, we perform uniformly sampling for a node to build the samples set as in \cite{HamiltonYL17}, where the number of sampled nodes in each layer is set as 5 and 3 across all datasets. In addition, half of the nodes are sampled for capturing macro-dynamics and micro-dynamics, respectively.

For the LIF model, we fix $\tau_m=1.0$ for all datasets. We empirically set the membrane reset potential $V_\text{reset}=0$, firing potential $V_\text{th}=1.0$ and threshold decays $\tau_\text{th}=0.7$, $\gamma=0.2$ for all datasets. To facilitate surrogate gradient learning, the smooth Sigmoid function is used as a threshold function to approximate the gradients during backpropagation with smooth factor $\alpha=1.0$ across all datasets.
All models are implemented with PyTorch~\cite{pytorch}. All our experiments are conducted on one NVIDIA TITAN RTX GPU with 24GB memory.

\subsection{Additional Experimental Results}
\subsubsection{Firing sparsity}
\begin{figure}[t]
    \centering
    \includegraphics[width=0.5\linewidth]{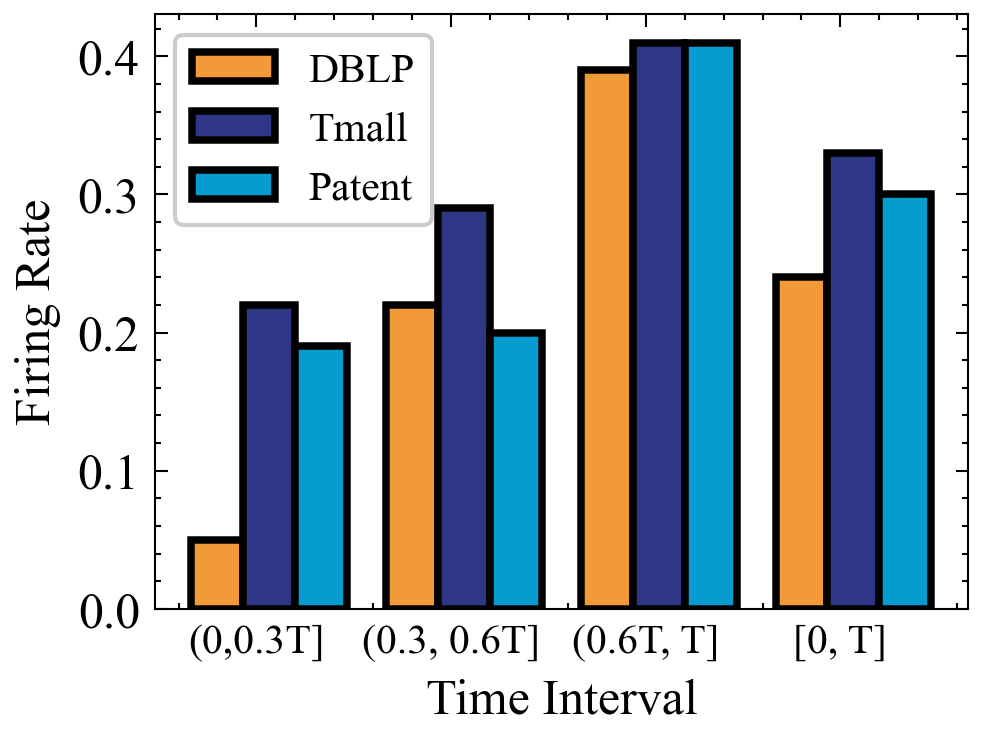}
    \caption{The average firing rate of SpikeNet at different time intervals. }
    \label{fig:firing_rate}
\end{figure}

Benefiting from intrinsic neuronal dynamics and spike-based communication paradigms, SNNs can be easily applied on some specialized neuromorphic hardware, such as Intel Loihi~\cite{davies2018loihi} or Brainchip Akida~\cite{vanarse2019hardware}. As for the power-efficient neuromorphic computation, the firing rate is an important property for SNNs since the energy consumption is proportional to the number of spikes \cite{patel2021spiking}. Reduction of firing rate is essential for energy-efficient neuromorphic chips. To this end, we hereby calculate the average firing rate of SpikeNet in the intermediate representations at a different time interval and show the results in Figure~\ref{fig:firing_rate}. Specifically, the firing rate is calculated by $\#spikes/(\#time\ steps \times \#neurons)$. A neuron with a high firing rate will lose the advantages of temporal sparsity and harm the effective representation in a temporal graph.
In DBLP, SpikeNet has a lower firing rate, especially in the prior time steps. The result suggests that the evolutionary pattern of nodes and edges in DBLP in short term is not significant, which leads to a sparser response compared with Tmall and Patent. In addition, the results also show that the intermediate node representations of SpikeNet are much sparse (only 20\% to 30\% of elements are non-zero). In other words, we need only about 20\% to 30\% of the memory to store the intermediate node representations. Moreover, each dimension of the embedding vectors learned by SpikeNet is only encoded by 1 bit (binary spikes), different from the real-valued embedding vectors that are encoded by at least 32 bits. As a result, the binarized representations can also greatly reduce the memory and time cost for the downstream tasks.

\end{document}